\documentclass{article}
\usepackage{spconf,amsmath,amssymb,graphicx,hyperref}
\usepackage{multirow}

\title{Bridge-TS: Exploiting the Prior of Generative Time Series Imputation}
\begin{document}

 \name{Yuyang Miao\sthanks{Thanks to XYZ agency for funding.}, Chang Li\sthanks{Co-first Author}, Zehua Chen\sthanks{Corresponding Author}}
\address{Imperial College London, Department of Electronic and Electrical Engineering, London, UK \\
University of Science and Technology of China, Hefei China \\
Tsinghua University, Department of CST, Beijing, 100084, China \\
Shengshu AI, Beijing, China}

        
%
\maketitle
\begin{abstract}
  Time series imputation, \textit{i.e.}, filling the missing values of a time recording, finds various applications in electricity, finance, and weather modelling. 
Previous methods have introduced generative models such as diffusion probabilistic models and Schrodinger bridge models to conditionally generate the missing values from Gaussian noise or directly from linear interpolation results.
However, as their prior is not informative to the ground-truth target, their generation process inevitably suffer increased burden and limited imputation accuracy.
In this work, we present Bridge-TS, building a data-to-data generation process for generative time series imputation and exploiting the design of prior with two novel designs. 
Firstly, we propose \textit{expert prior}, leveraging a pretrained transformer-based module as an expert to fill the missing values with a deterministic estimation, and then taking the results as the prior of ground truth target.
Secondly, we explore \textit{compositional priors}, utilizing several pretrained models to provide different estimation results, and then combining them in the data-to-data generation process to achieve a compositional priors-to-target imputation process.
Experiments conducted on several benchmark datasets such as ETT, Exchange, and Weather show that Bridge-TS reaches a new record of imputation accuracy in terms of mean square error and mean absolute error, demonstrating the superiority of improving prior for generative time series imputation.   

\end{abstract}





\maketitle

\section{Introduction}
\begin{figure}
\centering
\includegraphics[width=\linewidth]{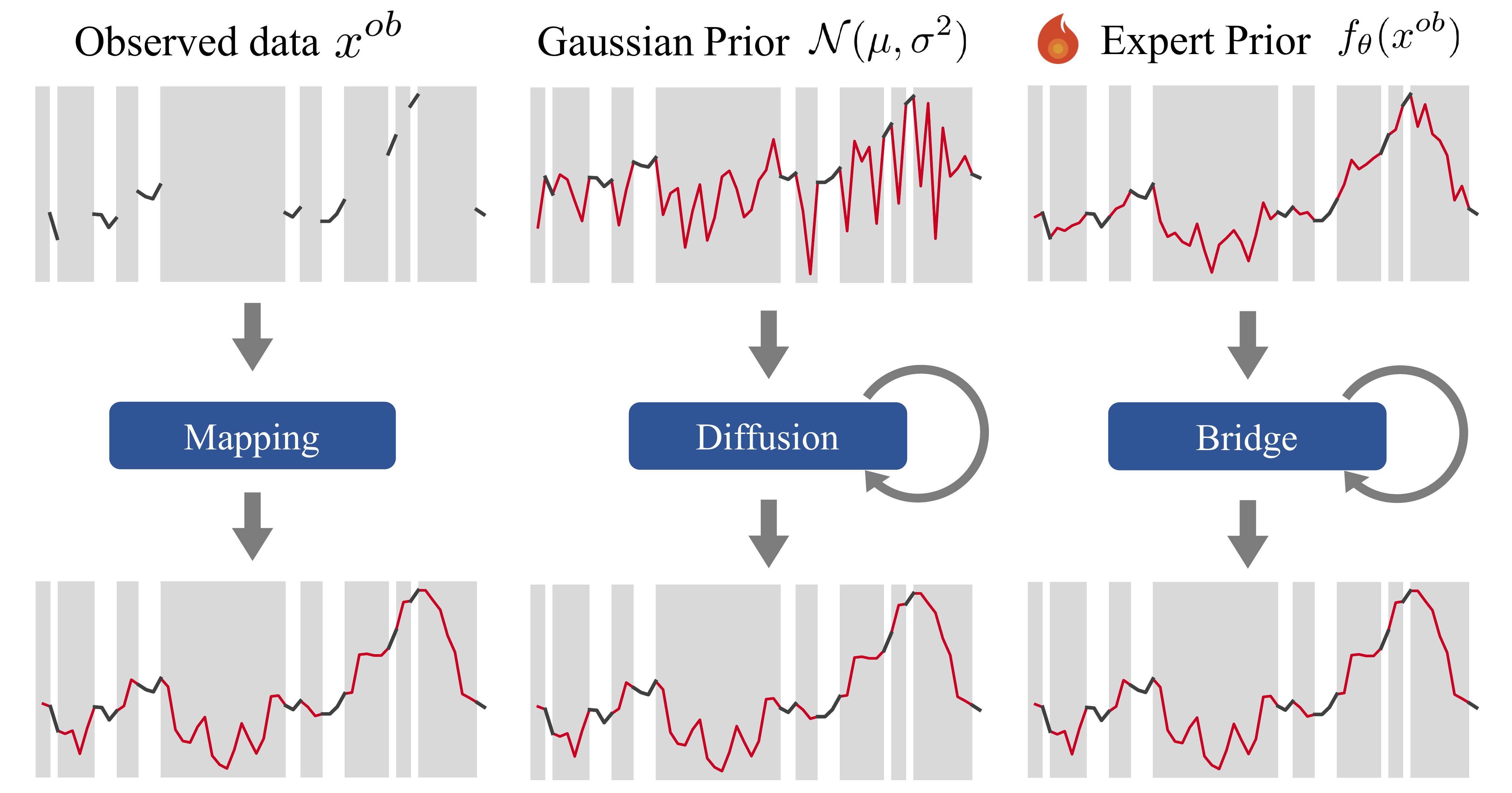} 
\caption{Comparison of three time series imputation approaches. Deterministic methods directly map observed data to targets, ignoring missing segments' priors. Diffusion models use iterative Gaussian sampling and our proposed Bridge-TS starts with expert priors, providing a more precise starting point and efficiently reaching the target distribution via bridge models.}
\label{fig:figure1}
\end{figure}

Time series data are foundational across a wide range of real-world applications, including electricity demand forecasting, financial trend analysis, and climate modeling. However, missing values frequently occur due to sensor failures, irregular sampling, or communication noise. These missing segments significantly degrade the performance of downstream tasks, making time series imputation a critical problem in data-driven modeling pipelines.

Recently, deep generative models have emerged as a promising direction for time series imputation, owing to their ability to capture uncertainty and complex temporal dynamics. Among them, denoising diffusion probabilistic models (DDPMs)\cite{kong2020diffwave, yuan2024diffusion} and Schrödinger Bridge (SB) models\cite{chen2023schrodinger, wang2024framebridge} have achieved state-of-the-art results in a variety of sequence modeling tasks. These methods formulate imputation as a conditional generation process, progressively denoising a prior distribution (e.g., Gaussian noise or linear interpolation) toward the missing target values. However, their performance remains bottlenecked by the choice of prior, which is often non-informative or statistically distant from the ground truth. As a result, such models may fail to reconstruct high-quality patterns in real-world time series.

Despite their recent success, generative models for time series imputation still face a fundamental limitation: the gap between the chosen prior distribution and the target data distribution. Most existing diffusion or bridge-based methods start from an uninformed prior, such as Gaussian noise or linear interpolation, which is often statistically far from the ground truth. As a result, these models may struggle to reach optimal solutions

On the other hand, deterministic models—such as TimesNet~\cite{wu2022timesnet}, Non-stationary Transformer~\cite{liu2022non}, and FEDformer~\cite{zhou2022fedformer}—achieve high point-wise accuracy by leveraging domain-specific architectural designs and tailored optimization objectives. 
However, these models lack the generative flexibility and uncertainty modeling capacity inherent to probabilistic frameworks. 

Therefore, the central challenge addressed in this paper is to design better priors for generative models. Our goal is to combine the advantage between deterministic accuracy and generative expressiveness by providing meaningful, data-driven priors that guide the generation process more effectively.

To this end, we introduce \textbf{Bridge-TS}, a novel Bridge-based generative model for time series imputation, which leverages expert knowledge from deterministic models to construct informative priors.

First, we introduce the concept of an \textit{expert prior}, which uses the output of a pretrained deterministic model as the prior for Schrödinger Bridge . This expert prior directly estimates missing segments based on observed values, significantly reducing the information gap between the prior and the target. The SB model is then trained to refine this prior toward the true data distribution, leading to faster convergence and higher imputation fidelity.

Second, we propose \textit{compositional priors}, where multiple deterministic models are used jointly to produce comprehensive and diverse prior estimates. These compositional priors are concatenated and refined jointly through the SB model. This compositional strategy enhances robustness and allows the generative model to benefit from the complementary strengths of different deterministic estimators.

Theoretically, our method leverages a closed-form solution of the Schrödinger Bridge under the linear-Gaussian assumption, allowing tractable simulation of the forward and backward marginal distributions (Eq.\ref{pt}). During training, the SB model learns to map time-indexed latent representations toward the target distribution using a simple L2 loss (Eq.\ref{bridgeloss}, \ref{bridgeloss}), resulting in a direct, data-to-data generative trajectory from the prior to the ground truth.

In summary, our contributions are as follows:

\begin{itemize} \item We propose \textbf{Bridge-TS}, a Schrödinger Bridge-based generative model that introduces data-driven priors for time series imputation, bridging the gap between deterministic estimation and generative modeling. \item We design and validate the use of \textit{expert prior} which reduces the burden on the SB model and improves imputation quality. \item We extend this to \textit{compositional priors}, where multiple \textit{expert priors} are fused within the SB framework, enabling more robust and accurate imputation. \item Experiments on six benchmark datasets—including ETT, Exchange, and Weather—demonstrate that Bridge-TS achieves state-of-the-art imputation accuracy in terms of MSE and MAE. \end{itemize}

\section{Related Works}

\subsection{Signal Processing Methods}  
Traditional time series imputation approaches are grounded in statistical signal processing, such as Wiener filtering~\cite{wiener1949extrapolation}, autoregressive modeling~\cite{bashir2018handling}, and Kalman filtering~\cite{durbin2012time}. 
While efficient and interpretable, these methods are limited in their ability to model nonlinearity, non-stationarity, and long-range dependencies, which are prevalent in real-world multivariate time series. Moreover, they are sensitive to gap length and noise, and lack adaptability across datasets. Their performance often deteriorates under high missing ratios or in dynamic environments, making them less suitable as standalone solutions for modern imputation tasks.

\subsection{Deterministic Deep Learning Methods}  
Deterministic deep learning methods have become dominant in time series modeling due to their capacity to learn complex temporal dependencies directly from data. Among them:

\textbf{TimesNet}~\cite{wu2022timesnet} proposes a temporal 2D-variation modeling framework that explicitly utilizes periodic structures in time series. Instead of processing raw 1D sequences, TimesNet identifies dominant periods and reshapes the input into 2D tensors to separate inter-period and intra-period variations. Within each TimeBlock, Inception-style convolutional modules~\cite{szegedy2015going} extract features across multiple receptive fields. These features are then aggregated and projected back to 1D for prediction. 

\textbf{FEDformer}~\cite{zhou2022fedformer} (Frequency Enhanced Decomposed Transformer) incorporates a seasonal-trend decomposition strategy and applies attention in the frequency domain. By leveraging the sparsity of time series in the Fourier space, FEDformer reduces computational cost and improves the model's capacity to extract long-range dependencies. Its frequency-aware design is especially beneficial for datasets with strong cyclic patterns.

\textbf{Non-stationary Transformer}~\cite{liu2022non} tackles the issue of non-stationarity by learning dynamic de-stationary transformations. It predicts scaling and shifting parameters through two MLPs, which are used to guide the attention learned to what it should be for a non-stationarized time series. This enables the model to retain the non-stationary properties of the original data while improving the stability of attention learning.

Other deterministic models also contribute valuable design insights. Autoformer~\cite{wu2021autoformer} introduces trend-seasonality decomposition and a progressive decomposition architecture, which facilitates the extraction of interpretable components. DLinear~\cite{zeng2023transformers} proposes a simple yet effective linear mapping model that surprisingly outperforms many Transformer-based models in some short-term settings, highlighting the importance of simplicity and inductive bias. iTransformer~\cite{liu2023itransformer} improves upon the vanilla Transformer by enhancing input encoding and interaction mechanisms. 

While these deterministic models provide accurate point estimates and capture various structural aspects of time series, they are inherently limited in handling uncertainty, generating diverse outcomes, or adapting flexibly to ambiguous imputations. In our work, we treat these models—particularly TimesNet, FEDformer, and Non-stationary Transformer—as informative expert priors. Bridge-TS uses their predictions as initialization and further refines them via a probabilistic Schrödinger Bridge, effectively combining deterministic strength with generative flexibility.

\subsection{Generative models based methods.}
Generative models have also been introduced to the field of time series imputation. Luo \textit{et al.} applied GAN in time series imputation \cite{luo2018multivariate}. Fortuin et al. utilized VAE as the imputation backbone \cite{fortuin2020gp}. CSDI \cite{tashiro2021csdi} reformulated the time series imputation task as a conditional generation task. Structured state space models are also applied for time series imputation \cite{alcaraz2022diffusion} .

\section{Preliminaries}
\subsection{Problem statement}
The task of time series imputation aims to complete the missing values $x^{ta}$ in the time series, given partial observations $x^{ob}$. According to the concepts introduced in CSDI \cite{tashiro2021csdi} and PulseDiff \cite{jenkins2023improving}, a time series can be defined as $X = \{x_{l,c}\} \in \mathbb{R}^{L \times C}$, where $L$ is the length of the time series and $C$ is the number of features. The observation mask is defined as $M = \{m_{l,c}\}\in\{0, 1\}^{L \times C}$, where $m_{l,c}=0$ indicates that $x_{l,c}$ is missing, and $m_{l,c}=1$ indicates that $x_{l,c}$ is observed. The goal of generative time series imputation is to predict the probability distribution of the missing portion \(p(x^{ta} \! \mid\! x_{\text{prior}}^{ob})
\) based on prior knowledge of the missing part $x_{\text{prior}}^{ta} = f_{\text{prior}}(x^{ob})$ when the observed values in $X$ are given. Figure \ref{fig:figure1} illustrates our system, Bridge-TS, which optimizes time series imputation from the dual perspectives of sample quality and probability distribution loss, guided by the power of compositional priors. The following sections progressively introduce the tractable Schrödinger bridge, the core process of Bridge-TS, as well as the construction of expert priors and the further enhancement of the model's performance with compositional priors.

During training, we initially derive $x_t$ via interpolation between $x_0$ and $x_T$ with uniformly sampled $t$. By directly estimating $x_0$, $\nabla \log p_t$, or $\nabla \log \hat{\Psi}_t$, we can obtain an effective score estimator for the reverse SDE, which facilitates the transformation between the distributions of paired data.
\subsection{Schrödinger Bridge in paired data}
In score-based generative models (SGMs) \cite{song2020score}, a forward stochastic differential equation (SDE) is defined between the paired data:
\begin{equation}
    dx_t = f(x_t, t)\, dt + g(t)\, dw_t ,\quad x_0 \sim p_\text{data}, \; x_T \sim p_\text{prior}
    \label{SDE}
\end{equation}
where \(t \in [0, T]\) represents the continuous time index, and \(x_t\) denotes the state of the data at time \(t\). The drift is given by the vector field \(f(x_t, t)\), the diffusion by the scalar function \(g(t)\), and \(w_t\) is the standard Wiener process.

Subjected to the boundary conditions, the Schrödinger Bridge (SB) problem \cite{schrodinger1932theorie, chen2023schrodinger} seeks to find a path measure \(p\) that minimizes the Kullback-Leibler (KL) divergence \(\min_{p \in \mathcal{P}_{[0,T]}} D_{\mathrm{KL}}(p \, \| \, p^\text{ref})
    \quad \text{s.t.} \quad p_0 = p_\text{data}, \; p_T = p_\text{prior}\) from a reference path measure \(p^\text{ref}\), which describes the transition between \(p_\text{prior}\) and \(p_\text{data}\) defined by \eqref{SDE}, 
where \(\mathcal{P}_{[0,T]}\) denotes the collection of all path measures over the time interval \([0, T]\). In SB's theory \cite{wang2021deep, chen2021likelihood}, such a problem is equivalent to a couple of forward-backward SDEs with \(x_0 \sim p_\text{data}, x_T \sim p_\text{prior}\):
\begin{equation}
    \left\{\begin{array}{l}
dx_t = \left[ f(x_t, t) + g^2(t) \nabla \log \Psi_t(x_t) \right] dt + g(t) dw_t\\
dx_t = \left[ f(x_t, t) - g^2(t) \nabla \log \hat{\Psi}_t(x_t) \right] dt + g(t) d\bar{w}_t  
\end{array}\right.
\end{equation}
where the non-linear drift $\nabla \log \Psi_t(x_t)$ and $\nabla \log \hat{\Psi}_t(x_t)$ can be described by coupled partial differential equations (PDEs):
\begin{equation}
\left\{\begin{array}{l}
\frac{\partial\Psi}{\partial t}=-\nabla_{x}\Psi^{\top} f-\frac{1}{2}\operatorname{Tr}\left(g^{2}\nabla_{x}^{2}\Psi\right)\\
\frac{\partial\widehat{\Psi}}{\partial t}=-\nabla_{x}\cdot(\widehat{\Psi} f)+\frac{1}{2}\operatorname{Tr}\left(g^{2}\nabla_{x}^{2}\widehat{\Psi}\right)
\end{array}\right.
\end{equation}
\begin{subequations}
\begin{align}
\text{s.t.} \quad \Psi_0 \hat{\Psi}_0\!=\!p_{\text{data}}, \Psi_T \hat{\Psi}_T\!=\!p_{\text{prior}}, x_t \sim p_t\!=\!\Psi_t \hat{\Psi}_t
\end{align}
\end{subequations}

Solving such SB problem usually requires iterative optimization to simulate the stochastic processes \cite{liu20232, de2021diffusion}, and the simulation path is not fully tractable \cite{somnath2023aligned}, which limits its scalability and applicability.
\begin{figure*}[t]
\centering
\includegraphics[width=\textwidth]{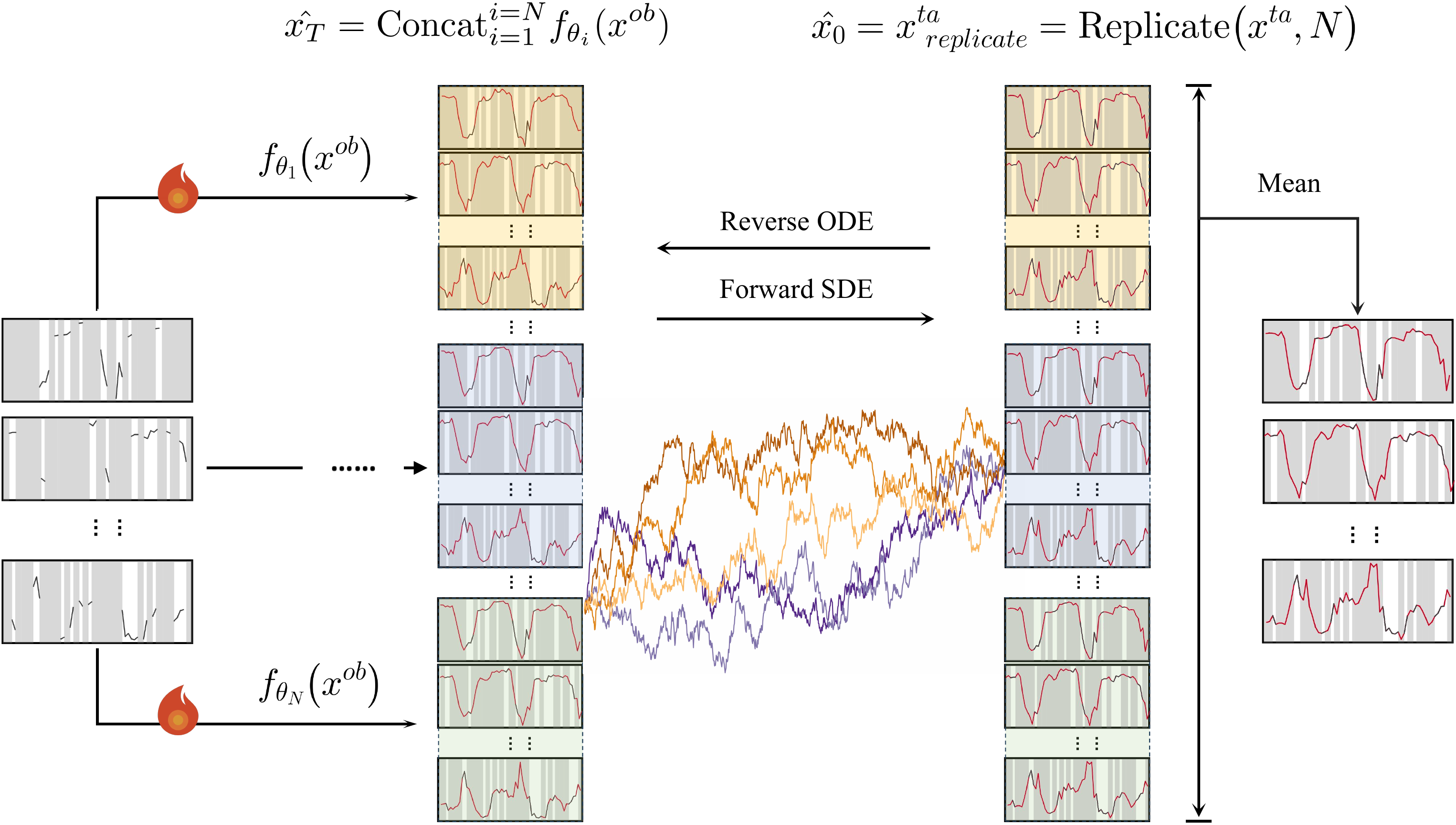} 
\caption{An example of how Bridge-TS works. Prior from different experts are concatenated. The concatenated priors are fed to the Schrödinger bridge with optimization trget as concatenated truth. The final output of the bridge are averaged channel-wise to give the final result. } 
\label{fig:figure1} 
\end{figure*}
\section{Methods}
\subsection{Bridge-TS}
Given the prior data $x_T = x_{\text{prior}}^{ta}$ in the missing segment, we utilize the SB model $p_{\theta}$ to generatively refine it towards the target distribution $x_0 = x^{ta} = p_{\theta}(x_{\text{prior}}^{ta})$.
A closed-form solution for SB \cite{chen2023schrodinger} exists when Gaussian smoothing is applied with $p_0 = \mathcal{N}(x_0, \epsilon_0^2 I)$ and $p_T = \mathcal{N}(x_T, \epsilon_T^2 I)$ to the original Dirac distribution. Under \(\epsilon_T = e^{\int_0^T f(\tau) d\tau} \epsilon_0\) with \(\epsilon_0 \to 0\), 
\(
\alpha_t = e^{\int_0^t f(\tau) d\tau}, \bar{\alpha}_t = e^{\int_t^1 f(\tau) d\tau},\sigma_t^2 = \int_0^t \frac{g^2(\tau)}{\alpha_\tau^2} d\tau,
\bar{\sigma}_t^2 = \int_t^1 \frac{g^2(\tau)}{\bar{\alpha}_\tau^2} d\tau
\), with boundary conditions and the linear Gaussian assumption, the tractable form of SB can be solved as
\begin{equation}
\hat{\Psi}_t = \mathcal{N}(\alpha_t x, \alpha_t^2 \sigma_t^2 I), \quad \Psi_t = \mathcal{N}(\bar{\alpha}_t y, \alpha_t^2 \bar{\sigma}_t^2 I)
\end{equation}
Therefore, the marginal distribution of \(x_t\) at \(t\) is also Gaussian and has a tractable form:

\begin{equation}
    \footnotesize
    x_t \sim p_t(x_0, x_T, f, g) = \Psi_t \hat{\Psi}_t = \mathcal{N} \left( \frac{\alpha_t \bar{\sigma}_t^2}{\sigma_1^2} x_0 + \frac{\bar{\alpha}_t \sigma_t^2}{\sigma_1^2} x_T, \frac{\alpha_t^2 \bar{\sigma}_t^2 \sigma_t^2}{\sigma_1^2} I \right).
    \label{pt}
\end{equation}

In contrast to diffusion models \cite{kong2020diffwave, ho2020denoising}, which assume a Gaussian prior distribution and construct a lengthy noise-to-data trajectory, employing either training-based conditional sampling \cite{tashiro2021csdi} or training-free sampling \cite{yuan2024diffusion} for imputation, Bridge-TS directly establishes a data-to-data evolution through predicting the original data at different timesteps, which allows the simulation of the SB process from \(t = T\) to \(t = 0\):
\begin{equation}
    \footnotesize
    \mathcal{L}_{\text{bridge}} = \mathbb{E}_{x^{ta} \sim p_{\text{data}}, x_\text{prior}^{ta} \sim p_{\text{prior}}, M} \mathbb{E}_t \left[ \left\| p_{\theta}(x_t, t, x_\text{prior}^{ta}, M) - x^{ta} \right\|_2^2 \right]
    \label{bridgeloss}
\end{equation}
This approach is advantageous for fully leveraging the informative content from the prior data and significantly reduces the number of sampling steps, all while harnessing the strong distribution fitting capability of the generative model.

\subsection{Expert Prior}
In concurrent work, some models utilize Digital Signal Processing (DSP)-based methods to provide priors for the missing parts \cite{park2024leveraging}. However, such priors constructed based on simple heuristic methods often fail to capture the nuanced data distribution or effectively utilize observed information \(x^{ob}\), resulting in low accuracy and making them unsuitable as optimal priors. 

On the other hand, certain deterministic methods \cite{wu2022timesnet, liu2022non}, which directly target the optimization of Mean-Square-Error (MSE) , have demonstrated remarkable performance in specific tasks. To maximize the informativeness of our prior and establish a more advantageous starting point for our model, we strive to reduce the divergence between the prior and the actual data.
Consequently, constructing our model directly upon these expert priors for optimization is anticipated to yield superior initial conditions for the SB process. Specifically, considering the time series dataset \(\mathcal{D}_{\text{train}}, \mathcal{D}_{\text{val}}, \mathcal{D}_{\text{test}} = \{X, M\}\), we define \(f_\theta\) as the expert prior that minimizes the validation error on \(\mathcal{D}_{\text{val}}\). Accordingly, we obtain:
\begin{equation}
x_\text{prior}^{ta} = f_\theta(x^{ob}), \quad \text{where } f_\theta \in \arg\min_{f} \text{MSE}(f(x^{ob}); \mathcal{D}_{\text{val}})
\end{equation}

\subsection{Compositional Priors}
Multiple works have been proposed to provide deterministic imputation results in time series tasks, each demonstrating robustness and performance. Optimizing from a single prior perspective may sometimes result in suboptimal predictions or failures to fully exploit the complexity of the data. Thus, an interesting question arises: Can the combined power of multiple priors lead to further improvements in the imputation task?

To explore this question, we experiment with a set of multiple priors, denoted as \( \{f_{\theta_i}\}_{i=1}^N \), where each \( f_{\theta_i} \) represents an expert prior trained under different assumptions or models. These priors are combined within our framework to influence the imputation process jointly.
At this stage, the SB model not only serves as a refinement process from prior to data but also requires the capability to effectively fuse and filter multiple priors.

Given the observed data \(x^{ob}\), each prior \(f_{\theta_i}, i \in [1, N]\), is applied to generate predictions for the missing values, which are concatenated along the channel dimension:
\begin{equation}
\hat{x_T} = \text{Concat}_{i=1}^{i=N}f_{\theta_i}(x^{ob}) \in \mathbb{R}^{L \times C \times N},
\end{equation}
Simultaneously, the ground truth missing part \(x^{ta}\) is replicated along the channel dimension to serve as \(x_0\) of the SB model, and we can get \(\hat{x_t}\) through \eqref{pt}:
\begin{equation}
\footnotesize
\hat{x_0} = x_{\text{replicate}}^{ta} = \text{Replicate}(x^{ta}, N) \in \mathbb{R}^{L \times C \times N}, \hat{x_t} = p_t(\hat{x_0}, \hat{x_T}, f, g)
\end{equation}
Both expert prior estimation model and bridge model are jointed trained, and the final prediction target is obtained through averaging in original channel, thus form the final training loss with \eqref{bridgeloss}:
\begin{equation}
\footnotesize
\mathcal{L}_{\text{final}} = \mathbb{E}_{x^{ta} \sim p_{\text{data}}, x^{ob} \sim p_{\text{prior}}, M} \mathbb{E}_t \left[ \left\| \frac{1}{N} \sum_{i=1}^{N} p_{\theta}(\hat{x_t}, t, \hat{x_T}, M)- x^{ta} \right\|_2^2 \right]
\end{equation}

\begin{figure*}
    \centering
    \includegraphics[width=1\linewidth]{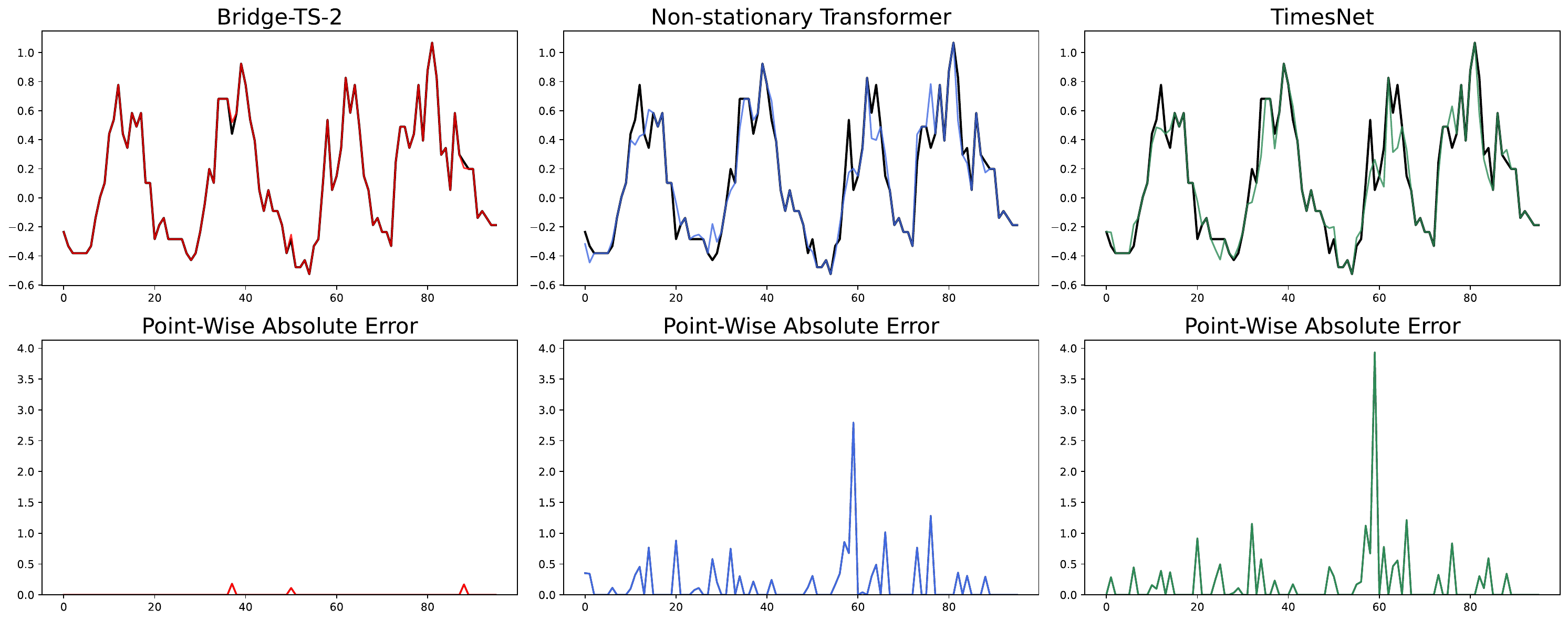}
    \caption{An example of Bridge-TS-2 results. Black line corresponds to the ground truth. Red, blue and green corresponds to the outputs of Bridge-TS-2, Non-stationary Transformer and TimesNet. Non-stationary Transformer and TimesNet are used as compositional priors for the Bridge-TS-2 model. The top row gives each model's imputation results and the bottom row gives point-wise imputation errors. It is obvious that Bridge-TS-2 can generate imputed time series that has smaller errors from compositional priors with larger errors.}
    \label{fig:example}
\end{figure*}
\section{Experiments}

To comprehensively evaluate the performance of our proposed model, we conducted extensive experiments across a diverse set of real-world datasets, ensuring a thorough assessment of its imputation capabilities under various conditions.

\subsection{Datasets} Time series data often suffer from missing values due to sensor malfunctions, data transmission errors, or other unforeseen issues. To simulate realistic scenarios, we evaluated our model on six publicly available datasets spanning electricity, weather, and finance domains:

\textbf{ETT (Electricity Transformer Temperature)}: This dataset measures the temperature of electricity transformers, a critical indicator of electricity consumption. It comprises four subsets:

\begin{itemize}
    \item \textbf{ETTh1} and \textbf{ETTh2}: Hourly sampled data over two years.
    \item \textbf{ETTm1} and \textbf{ETTm2}: 15-minute interval data over the same period.
\end{itemize}

Each subset contains seven features: one oil temperature and six power load indicators.

\textbf{Weather}: Collected every 10 minutes throughout 2022, this dataset includes 21 meteorological indicators such as temperature, humidity, and wind speed, providing a comprehensive view of weather patterns.

\textbf{Exchange}: This financial dataset records daily exchange rates from 1990 to 2016 for eight countries: Australia, the United Kingdom, Canada, Switzerland, China, Japan, New Zealand, and Singapore. It offers insights into long-term financial trends and currency fluctuations.

\section{Experiments}
To evaluate our model's performance, we have extensively conducted experiments on a wide range of datasets.

\subsection{Datasets} In real-world scenarios, time series signals often suffer from missing data due to malfunctions or transmission errors. Therefore, imputing missing values in time series is both common and essential. In this work, we assessed our model's imputation capability on six datasets spanning various domains: electricity \cite{zhou2021informer} (ETTh1, ETTh2, ETTm1, ETTm2), weather information (Weather) \cite{mpg_weather}, and financial data (Exchange) \cite{lai2018modeling}.

The Electricity Transformer Temperature (ETT) datasets measure the temperature of electricity transformers, a critical factor indicating electricity consumption. Two years of ETT data were collected and divided into different granularities. ETTh1 and ETTh2 contain data sampled at 1-hour intervals, while ETTm1 and ETTm2 are sampled at 15-minute intervals. Each dataset comprises seven channels: one for oil temperature and six for power loads.

The Weather dataset includes 21 dimensions of meteorological indicators, such as temperature and humidity. Data were collected every 10 minutes throughout 2022.

The Exchange dataset is eight-dimensional, recording the exchange rates of eight countries: Australia, the United Kingdom, Canada, Switzerland, China, Japan, New Zealand, and Singapore. The data span daily records from 1990 to 2016.

\subsection{Baselines} To ensure a robust comparison, we benchmarked our model against six state-of-the-art time series imputation methods:

\textbf{TimesNet} \cite{wu2022timesnet} is a CNN-based architecture that transforms 1D time series into 2D, leveraging inter-period and intra-period.

\textbf{Non-stationary Transformer} \cite{liu2022non} incorporates non-stationary components into the Transformer framework, allowing it to adapt to varying statistical properties over time.

\textbf{iTransformer} \cite{liu2023itransformer} enhances Transformer models by applying the attention mechanism and feed-forward network to the inverted dimensions.

\textbf{DLinear} \cite{zeng2023transformers} is a linear model tailored for efficient time series processing, emphasizing simplicity and speed.

\textbf{Autoformer} \cite{wu2021autoformer} utilizes decomposition blocks to separate trend and seasonal components, improving global information extraction.

\textbf{FEDformer} \cite{zhou2022fedformer} integrates frequency-enhanced decomposition techniques to capture both time and frequency domain information.

These baselines encompass a diverse range of architectures, including Transformer-based, convolutional, and linear models, providing a comprehensive evaluation landscape.

\subsection{Implementation Details} Following the methodology outlined in TimesNet, we introduced missing values into the datasets by randomly masking individual data points across all channels with probabilities of 12.5\%, 25.0\%, 37.5\%, and 50.0\%. This approach simulates varying degrees of data incompleteness, reflecting real-world scenarios. The sequence length for all experiments was fixed at 96 time steps.

Given their consistent superior validation performance and complementary characteristics, TimesNet and Non-stationary Transformer were selected as compositional priors in our model. Specifically, TimesNet served as the backbone for the Schrödinger Bridge component due to its ability to effectively handle multichannel inputs in a non-compressive way.

We explored three configurations of our proposed model:

\textbf{Bridge-TS-1} incorporates a single expert prior, either TimesNet or Non-stationary Transformer, chosen based on validation performance.

\textbf{Bridge-TS-2} employs both TimesNet and Non-stationary Transformer as compositional priors, leveraging their complementary strengths.

\textbf{Bridge-TS-3} integrates an additional expert prior, FEDformer, to assess the impact of increasing the number of compositional priors on imputation performance.

All expert priors were pretrained on their respective datasets and subsequently fine-tuned jointly with the Schrödinger Bridge model using the bridge loss function.

\subsection{Evaluation Metrics} To quantitatively assess the imputation performance, we utilized two widely adopted error metrics:

\textbf{Mean Squared Error (MSE)} is defined as $\text{MSE} = \frac{1}{n}\sum_{i=1}^{n}(y_i - x_{0_i})^2$, where $y_i$ represents the imputed value and $x_{0_i}$ denotes the ground truth. MSE penalizes larger errors more heavily, making it sensitive to outliers.

\textbf{Mean Absolute Error (MAE)} is calculated as $\text{MAE} = \frac{1}{n}\sum_{i=1}^{n}|y_i - x_{0_i}|$, offering a more robust measure by treating all error magnitudes linearly.

These metrics provide a comprehensive evaluation of imputation accuracy, with lower values indicating superior performance.

\subsection{Results}

\subsection{Comparison with Baseline Methods.} Table~\ref{Results} reports the performance of our proposed Bridge-TS model against a range of competitive baselines. Our model is evaluated on six real-world datasets under four different missing ratios (12.5\%–50\%).

Across all datasets and mask ratios, Bridge-TS consistently achieves the lowest or on-par MSE and MAE compared to all baselines, demonstrating strong generalization across diverse temporal structures and imputation difficulties. Notably, Bridge-TS outperforms the strongest deterministic models such as TimesNet and Non-stationary Transformer on the majority of tasks, suggesting that the generative refinement enabled by Schrödinger Bridge further boosts the performance beyond deterministic estimation alone. These results validate the effectiveness of our data-to-data generative formulation and the importance of incorporating accurate priors into the generative process.

\setlength{\tabcolsep}{3pt} 
\renewcommand{\arraystretch}{1.5} 

\begin{table*}[htbp]
\centering
\scriptsize 
\begin{tabular}{p{0.25cm}|p{0.70cm}|cc|cc|cc|cc|cc|cc|cc}
\hline
\multirow{2}{*}{\rotatebox{90}{DS}} & \multirow{2}{*}{Ratio} & \multicolumn{2}{c|}{Bridge-TS-2} & \multicolumn{2}{c|}{TimesNet} & \multicolumn{2}{c|}{Non-stationary} & \multicolumn{2}{c|}{iTransformer} & \multicolumn{2}{c|}{Dlinear} & \multicolumn{2}{c|}{AutoFormer} & \multicolumn{2}{c}{FEDformer} \\
& & \multicolumn{1}{c}{MSE} & \multicolumn{1}{c|}{MAE} & \multicolumn{1}{c}{MSE} & \multicolumn{1}{c|}{MAE} & \multicolumn{1}{c}{MSE} & \multicolumn{1}{c|}{MAE} & \multicolumn{1}{c}{MSE} & \multicolumn{1}{c|}{MAE} & \multicolumn{1}{c}{MSE} & \multicolumn{1}{c|}{MAE} & \multicolumn{1}{c}{MSE} & \multicolumn{1}{c|}{MAE} & \multicolumn{1}{c}{MSE} & \multicolumn{1}{c}{MAE} \\\hline 
\multirow{4}{*}{\rotatebox{90}{ETTh1}}  & 12.5\% &\textbf{0.037} &\textbf{0.131} &0.061 &0.168 &0.037\textsuperscript{*} &0.131\textsuperscript{*} &0.132 &0.254 &0.113 &0.233 &0.092 &0.222 &0.062 &0.180 \\
                        & 25.0\% &\textbf{0.048} & \textbf{0.146} &0.086 &0.229 &0.051 &0.150 &0.125 &0.349 &0.149 &0.269 &0.127 &0.259 &0.089 & 0.217\\
                        & 37.5\% &\textbf{0.061} &\textbf{0.165} &0.102 &0.213 &0.068 &0.177 &0.384 &0.453 &0.188 &0.302 &0.153 & 0.281 &0.127 &0.260\\
                        & 50.0\%&\textbf{0.080} &\textbf{0.189} &0.121 &0.230 &0.093 &0.205 &0.510 &0.500 &0.230 &0.332 &0.213 &0.334 &0.176 &0.309\\ 
                        \hline
\multirow{4}{*}{\rotatebox{90}{ETTh2}}  & 12.5\% &\textbf{0.034} &\textbf{0.116} &0.041 &0.135 &0.036 &0.117 &0.161 &0.271 &0.114 &0.226 &0.174 &0.297 &0.097 &0.213 \\
                        & 25.0\% &\textbf{0.037} & \textbf{0.122} &0.049 &0.148 &0.041 &0.127 &0.415 &0.442 &0.151 &0.263 &0.206 &0.316 &0.137 & 0.258\\
                        & 37.5\% &\textbf{0.042} &\textbf{0.131} &0.055 &0.157 &0.044 &0.134 &0.877 &0.626 &0.185 &0.294 &0.280 & 0.356 &0.193 &0.307\\
                        & 50.0\%&\textbf{0.048} &\textbf{0.132} &0.062 &0.166 &0.050 &0.146 &1.628 &0.808 &0.220 &0.322 &0.352 &0.409 &0.308 &0.382\\
                        \hline
\multirow{4}{*}{\rotatebox{90}{ETTm1}}  & 12.5\% &\textbf{0.016} &\textbf{0.084} &0.018 &0.089 &0.017 &0.087 &0.067 &0.181 &0.058 &0.164 &1.089 &0.873 &0.034 &0.131 \\
                        & 25.0\% &\textbf{0.019} & \textbf{0.092} &0.022 &0.097 &0.019\textsuperscript{*} &0.093 &0.140 &0.268 &0.080 &0.194 &0.685 &0.663 &0.052 & 0.162\\
                        & 37.5\% &\textbf{0.024} &\textbf{0.101} &0.027 &0.108 &0.024\textsuperscript{*} &0.101\textsuperscript{*} &0.238 &0.346 &0.104 &0.222 &0.408 & 0.479 &0.080 &0.199\\
                        & 50.0\%&\textbf{0.029} &\textbf{0.112} &0.033 &0.118 &0.029\textsuperscript{*} &0.112\textsuperscript{*} &0.329 &0.399 &0.133 &0.250 &0.329 &0.415 &0.125 &0.252\\
                        \hline
\multirow{4}{*}{\rotatebox{90}{ETTm2}}  & 12.5\% &\textbf{0.015} &\textbf{0.071} &0.018 &0.077 &0.016 &0.072 &0.113 &0.224 &0.070 &0.172 &0.977 &0.784 &0.060 &0.167 \\
                        & 25.0\% &\textbf{0.017} & \textbf{0.077} &0.019 &0.083 &0.018 &0.077 &0.348 &0.401 &0.0925 &0.203 &0.911 &0.727 &0.090 & 0.206\\
                        & 37.5\% &\textbf{0.020} &\textbf{0.083} &0.022 &0.090 &0.020\textsuperscript{*} &0.084 &0.806 &0.590 &0.114 &0.227 &0.683 & 0.608 &0.132 &0.247\\
                        & 50.0\%&\textbf{0.023} &\textbf{0.090} &0.024 &0.095 &0.023\textsuperscript{*} &0.090\textsuperscript{*} &1.590 &0.775 &0.140 &0.253 &0.654 &0.562 &0.240 &0.331\\
                        \hline
\multirow{4}{*}{\rotatebox{90}{Exchange}}  & 12.5\% &\textbf{0.0020} &\textbf{0.025} &0.0030 &0.031 &0.0033 &0.030 &0.0825 &0.2100 &0.0250 &0.108 &0.2127 &0.349 &0.1129 &0.221 \\
                        & 25.0\% &\textbf{0.0026} & \textbf{0.028} &0.0036 &0.035 &0.0041 &0.037 &0.037 &0.133 &0.033 &0.127 &0.2981 &0.390 &0.1896 & 0.2951\\
                        & 37.5\% &\textbf{0.0033} &\textbf{0.032} &0.0044 &0.039 &0.0047 &0.0418 &0.7998 &0.626 &0.0524 &0.158 &0.4569 & 0.471 &0.2778 &0.358\\
                        & 50.0\%&\textbf{0.0041} &\textbf{0.037} &0.0050 &0.043 &0.0055 &0.045 &1.655 &0.832 &0.0706 &0.184 &0.3860 &0.432 &5.452 &1.558\\
                        \hline
\multirow{4}{*}{\rotatebox{90}{Weather}}  & 12.5\% &\textbf{0.023} &\textbf{0.033} &0.027 &0.053 &0.026 &0.048 &0.056 &0.116 &0.042 &0.094 &0.265 &0.375 &0.042 &0.102 \\
                        & 25.0\% &\textbf{0.026} & \textbf{0.039} &0.029 &0.055 &0.029 &0.054 &0.115 &0.199 &0.053 &0.116 &0.256 &0.355 &0.060 & 0.138\\
                        & 37.5\% &\textbf{0.029} &\textbf{0.046} &0.033 &0.062 &0.033 &0.061 &0.214 &0.283 &0.059 &0.125 &0.130 & 0.237 &0.078 &0.165\\
                        & 50.0\%&\textbf{0.033} &\textbf{0.051} &0.035 &0.064 &0.039 &0.071 &0.365 &0.362 &0.069 &0.140 &0.141 &0.245 &0.114 &0.212\\
                        \hline\hline
\end{tabular}
\caption{Performance of Bridge-TS2 and other baselines for different datasets and mask ratios.}
\label{Results}
\end{table*}

\setlength{\tabcolsep}{3pt} 
\renewcommand{\arraystretch}{1.5} 
\begin{table*}[htbp]
\centering
\scriptsize 
\begin{tabular}{c|c|cc|cc|cc|cc|cc|cc|cc}
\hline
\multirow{2}{*}{DS} & \multirow{2}{*}{Ratio} & \multicolumn{2}{c|}{Linear-Bridge}& \multicolumn{2}{c|}{Bridge-TS-1} & \multicolumn{2}{c|}{Bridge-TS-2} & \multicolumn{2}{c|}{Bridge-TS-3} & \multicolumn{2}{c|}{TimesNet} & \multicolumn{2}{c|}{Non-stationary} & \multicolumn{2}{c}{FEDformer} \\

& & MSE & MAE & MSE & MAE & MSE & MAE & MSE & MAE & MSE & MAE & MSE & MAE & MSE & MAE\\
\hline

\multirow{4}{*}{\rotatebox{90}{ETTh1}} & 12.5\% & 0.049 & 0.145 & 0.038 & 0.130 & 0.037 & 0.131 & \textbf{0.036} & \textbf{0.128} & 0.061 & 0.168 & 0.037 & 0.131 & 0.062 & 0.180\\
                       & 25.0\% & 0.056 & 0.156 & 0.049 & 0.149 & 0.048 & 0.146 & \textbf{0.046} & \textbf{0.144} & 0.086 & 0.229 & 0.051 & 0.150 & 0.089 & 0.217 \\
                       & 37.5\% & 0.075 &  0.178& 0.063 & 0.168 & 0.061 & 0.165 & \textbf{0.059} & \textbf{0.163} & 0.102 & 0.213 & 0.068 & 0.177 & 0.127 & 0.260\\
                       & 50.0\% & 0.107& 0.206 & 0.082 & 0.192 & 0.080 & 0.189 & \textbf{0.079} & \textbf{0.188} & 0.121 & 0.230 & 0.093 & 0.295 & 0.176 & 0.309\\
                       \hline
                       
\multirow{4}{*}{\rotatebox{90}{ETTh2}} & 12.5\% &0.040 &0.125  &0.035 &0.120 &\textbf{0.034} &\textbf{0.116} &0.034 &0.117 &0.041 &0.135 &0.036 &0.117 &0.097 &0.213\\
                       & 25.0\% & 0.043& 0.129&0.039 &0.126 &\textbf{0.037} &\textbf{0.122} &0.038 &0.128 &0.049 &0.148 &0.041 &0.127 &0.137 &0.258\\
                       & 37.5\% & 0.049& 0.138&0.043 &0.134 &\textbf{0.042} &\textbf{0.131} &0.042 &0.135 &0.055 &0.157 &0.044 &0.134 &0.193 &0.307\\
                       & 50.0\% & 0.058& 0.150 &0.050 &0.145 &\textbf{0.048} &\textbf{0.132} &0.052 &0.152 &0.062 &0.166 &0.050 &0.146 &0.308 &0.382\\
                       \hline

\multirow{4}{*}{\rotatebox{90}{ETTm1}} & 12.5\% & 0.021& 0.097&0.019 &0.092 &\textbf{0.016} &\textbf{0.084} & 0.016\textsuperscript{*} & 0.085 &0.018 &0.089 &0.017 &0.087 &0.034 &0.131\\
                       & 25.0\% & 0.025& 0.103&0.022 &0.098 &\textbf{0.019} &\textbf{0.092} &0.020 &0.095 &0.022 &0.097 &0.019\textsuperscript{*} &0.093 &0.052 &0.162\\
                       & 37.5\% & 0.040& 0.118&0.028 &0.112 &\textbf{0.024} &\textbf{0.101} &0.025 &0.106 &0.027 &0.108 &0.024\textsuperscript{*} &0.101\textsuperscript{*} &0.080 &0.199\\
                       & 50.0\% &0.045 & 0.126 &0.0353 &0.1249 &\textbf{0.029} &\textbf{0.112} &0.030 &0.117 &0.033 &0.118 &0.029\textsuperscript{*} &0.112\textsuperscript{*} &0.125 &0.252\\
                       \hline 
\multirow{4}{*}{\rotatebox{90}{ETTm2}} & 12.5\% & 0.018& 0.077&0.016 &0.072 &\textbf{0.015} &\textbf{0.071} &0.016 &0.075 &0.018 &0.077 &0.016 &0.072 &0.060 &0.167\\       
                       & 25.0\% &0.025 &0.082 &0.018 &0.078 &\textbf{0.017} &\textbf{0.077} &0.018 &0.082 &0.019 &0.083 &0.018 &0.077\textsuperscript{*} &0.090 &0.206\\
                       & 37.5\% &0.028 &0.087 &0.021 &0.085 &\textbf{0.020} &\textbf{0.083} &0.021 &0.089 &0.022 &0.090 &0.020\textsuperscript{*} &0.084 &0.132 &0.247\\
                       & 50.0\% &0.033 &0.093 &0.024 &0.093 &\textbf{0.023} &\textbf{0.090} &0.024 &0.099 &0.024 &0.095 &0.023\textsuperscript{*} &0.090\textsuperscript{*} &0.240 &0.331\\
                       \hline
\multirow{4}{*}{\rotatebox{90}{Exchange}} & 12.5\% &\textbf{0.0020} &\textbf{0.020} &0.0023 &0.025 &0.0020 &0.025 &0.0027 &0.030 &0.0030 &0.031 &0.0033 &0.030 &0.1129 &0.221\\
                          & 25.0\% &\textbf{0.0020} &\textbf{0.022} & 0.0026 & 0.029 & 0.0026 & 0.028 & 0.0049 & 0.044 & 0.0036 & 0.035 & 0.0041 & 0.037 & 0.1896 & 0.2951\\
                          & 37.5\%  &\textbf{0.0025} &\textbf{0.024} & 0.0032 & 0.032 & 0.0029 & 0.029 & 0.0087 & 0.059 & 0.0044 & 0.039 & 0.0047 & 0.041 & 0.2778 & 0.358\\
                          & 50.0\% &\textbf{0.0031} &\textbf{0.027} & 0.0040 & 0.037 & 0.0041 & 0.037 & 0.019 & 0.098
                          & 0.0050 & 0.043 & 0.0055 & 0.045 & 5.452 & 1.558\\
                          \hline
\multirow{4}{*}{\rotatebox{90}{Weather}} & 12.5\% & 0.027& 0.042&0.024 &0.035 &\textbf{0.023} & \textbf{0.033} & 0.023\textsuperscript{*} & 0.034 &0.027 &0.053 & 0.026 & 0.048 &0.042 &0.102\\
                        &25.0\% & 0.030& 0.046&0.027 & 0.041 &0.026 &\textbf{0.039} & \textbf{0.025} & 0.041 & 0.029 & 0.055 &0.029 &0.054 & 0.060 & 0.138\\
                        &37.5\% & 0.032 & 0.033 & 0.031 & 0.049 & \textbf{0.029} & \textbf{0.046} & 0.029\textsuperscript{*} & 0.048 &0.033 &0.062 &0.033 &0.061 &0.078 & 0.165\\
                        &50.0\% &0.035 &0.036 & 0.035 & 0.057 & 0.033 & \textbf{0.051} & \textbf{0.032} & 0.055 & 0.035 & 0.064 &0.039 &0.071 &0.114 &0.212\\
                        \hline

\end{tabular}
\caption{The performance of Bridge-TS with linear prior and different compositional priors.}
\label{priors}
\end{table*}

\subsection{Effect of Expert Priors.} To evaluate the role of expert priors, we have also included a variant of our model—Bridge-TS-1—that uses a single expert prior. As shown in Table~\ref{priors}, Bridge-TS-1 consistently outperforms its corresponding expert prior across nearly all settings. This confirms that the Schrödinger Bridge can effectively refine the expert prediction and correct small temporal deviations, leading to more accurate imputation results.

In contrast, when using a simple linear interpolation prior (Linear-Bridge), performance drops significantly across most datasets, except for Exchange, where deterministic estimators are relatively weak. This supports our hypothesis that the informativeness of the prior plays a critical role in the effectiveness of the generative process: strong expert priors lead to better initialization and tighter refinement, whereas uninformed or noisy priors increase the burden on the generative model and hurt accuracy.

\subsection{Impact of Compositional Priors.} To further improve robustness, we extend our framework to compositional priors by integrating multiple expert estimators. Bridge-TS-2 utilizes two deterministic priors (TimesNet and Non-stationary Transformer), while Bridge-TS-3 incorporates an additional prior from FEDformer.

As shown in Table~\ref{priors}, Bridge-TS-2 achieves consistent gains over Bridge-TS-1 across all datasets and ratios, demonstrating that combining multiple priors enhances both the diversity and informativeness of the initial state. Interestingly, the benefit of adding a third prior (Bridge-TS-3) is conditional: for datasets where the third prior (e.g., FEDformer) performs well individually (such as ETTh1), Bridge-TS-3 yields further improvements. In contrast, when the additional prior is weak (e.g., FEDformer on Exchange), the fusion introduces noise and degrades performance. This phenomenon highlights an important insight: while compositional priors can enhance performance through model diversity, their quality and compatibility are essential. Including inaccurate priors can mislead the refinement process, emphasizing the need for reliable expert components in the compositional setting.

\subsection{Qualitative Case Study.} Figure~\ref{fig:example} illustrates a representative example from ETTh1, comparing the imputation results of Bridge-TS-2 with its underlying expert priors. While TimesNet and Non-stationary Transformer individually deviate from the ground truth in certain regions, Bridge-TS-2 successfully integrates their strengths and eliminates their biases, producing a more accurate reconstruction. The point-wise error plot further confirms that Bridge-TS achieves lower residuals across all timestamps, particularly in high-variation segments, validating its ability to extract and fuse useful signals from imperfect priors.

\begin{table}[h]
\centering
\scriptsize 
\begin{tabular}{c|cc|cc|cc|cc}
\hline
 \multirow{2}{*}{Ratio} & \multicolumn{2}{c|}{$f_{\theta}$}& \multicolumn{2}{c|}{$f_{\theta_{frozen}}$} & \multicolumn{2}{c|}{$f_{\theta_{MSE}}$} & \multicolumn{2}{c}{$f_{\theta_{MSE-N}}$} \\
&MSE &MAE &MSE &MAE &MSE &MAE &MSE &MAE\\
\hline
12.5\% & 0.037 & 0.131 & 0.048 & 0.143 & 0.048 & 0.144 & 0.267 & 0.338\\
25.0\% & 0.048 & 0.146 & 0.058 & 0.158 & 0.057 & 0.159 & 0.239 & 0.322\\
37.5\% & 0.061 & 0.165 & 0.068 & 0.171 & 0.072 & 0.177 & 0.374 & 0.414\\
50.0\% & 0.080 & 0.189 & 0.086 & 0.192 & 0.086 & 0.193 & 0.446 & 0.437\\
\hline
\end{tabular}
\caption{Abalation table on how the compositional priors should be jointly trained}
\label{ab2}
\end{table}

\begin{table}[h]
    \centering
    
    \begin{tabular}{c|c|c|c}
        \hline
        Steps &  $f_{\theta-N}$ & $f_{\theta_{frozen}}$ & $f_{\theta}$    \\
        \hline
        0 & 0.175 & 0.143 & 0.145\\
        \hline
        500 & 0.155 & 0.137 & 0.139\\
        \hline
        2000 & 0.141 & 0.121 & 0.109\\
        \hline
    \end{tabular}
    \caption{Traning dynamics of different compositional priors setup.}
    \label{ab3}
\end{table}

\setlength{\tabcolsep}{3pt} 
\renewcommand{\arraystretch}{1.5} 
\begin{table}[h]
\centering
\scriptsize 
\begin{tabular}{c|c|cc|cc|cc}
\hline

\multirow{2}{*}{Models} & \multirow{2}{*}{Mask Ratio} & \multicolumn{2}{c|}{Bridge-TS-2}& \multicolumn{2}{c|}{w/o Probabilstic} & \multicolumn{2}{c}{\(g_\text{constant}\)}   \\
&  &MSE &MAE &MSE &MAE &MSE &MAE \\
\hline

\multirow{4}{*}{ETTh1} & 12.5\% & 0.037 & 0.131 &0.054 &0.159 & 0.040& 0.136\\
                       & 25.0\% & 0.048 & 0.146 &0.082 &0.199 & 0.047& 0.147\\
                       &37.5\% &0.061 &0.165 &0.127 & 0.251 & 0.063& 0.168\\
                       &50.0\% &0.080 &0.189 &0.209 &0.328 & 0.079& 0.186\\
                       \hline                      
\multirow{4}{*}{ETTh2} & 12.5\% & 0.034 & 0.116 &0.049 &0.145 & 0.0356& 0.121\\
                       & 25.0\% & 0.037 & 0.122 &0.094 &0.207 & 0.038& 0.123\\
                       & 37.5\% &0.042 &0.131 &0.165 & 0.277 & 0.041& 0.013\\
                       & 50.0\% &0.048 &0.132 &0.487 &0.447 & 0.049& 0.146\\
                       \hline 
\multirow{4}{*}{ETTm1} & 12.5\% &0.016 &0.084 &0.027 & 0.111 & 0.016& 0.085\\
                       & 25.0\% &0.019 &0.092 &0.033 &0.124 & 0.020& 0.094\\
                       & 37.5\% &0.024 &0.101 &0.042 &0.140 & 0.025& 0.104\\
                       & 50.0\% &0.029 &0.112 &0.062 &0.171 & 0.031& 0.115\\
                       \hline
\multirow{4}{*}{ETTm2} & 12.5\% &0.015 &0.071 &0.023 &0.095 &0.015&0.070\\
                       & 25.0\% & 0.017 & 0.077 &0.034 &0.119 &0.017&0.076\\
                       & 37.5\% & 0.020 & 0.083 &0.054 & 0.150 &0.020&0.083\\
                       &50.0\% &0.023 &0.090 &0.109 &0.216 &0.023&0.090\\
                       \hline
\multirow{4}{*}{Exchange} & 12.5\% &0.0020 &0.025 &0.0030 &0.307 &0.0022&0.025\\
                          & 25.0\% &0.0026 &0.028 &0.0085 &0.058 &0.0029&0.028\\
                          & 37.5\% &0.0033 &0.032 &0.037 &0.135 &0.0032&0.031\\
                          & 50.0\% &0.0041 &0.037 &0.735 &0.633 &0.0040&0.036\\
                          \hline
\multirow{3}{*}{Weather} & 12.5\% &0.023 & 0.033 &0.024 &0.039 &0.023&0.032\\
                         & 25.0\% &0.026 &0.039 &0.027 &0.043 &0.026&0.038\\
                         & 37.5\% &0.029 &0.046 &0.035 &0.067 &0.029&0.045\\
                         & 50.0\% &0.033 &0.051 &0.045 &0.093 &0.032&0.051\\
                         \hline

\end{tabular}
\caption{Comparison of MSE and MAE across different models and mask ratios}
\label{ab}
\end{table}

\section{Ablation Studies}

We conduct extensive ablation studies to analyze the key components of Bridge-TS. Specifically, we investigate: (1) the necessity of generative modeling, (2) the impact of different noise schedules in the Schrödinger Bridge, and (3) strategies for training compositional priors.

\subsection{Effectiveness of Generative Modeling.}
To validate the necessity of generative modeling, we construct Bridge-TS-D, a variant of Bridge-TS-2 that uses a deterministic version of the Schrödinger Bridge module. This ensures a fair comparison as both models share identical architectures and parameters. As shown in Table~\ref{ab}, Bridge-TS-2 consistently outperforms Bridge-TS-D across all datasets and masking ratios. This confirms that the probabilistic formulation of the SB process enables stronger distributional modeling and refinement over deterministic cascades.

\subsection{Impact of Noise Schedules.}
We compare two noise schedules in the SB framework: the default constant schedule \( g_{\text{constant}} \) and the adaptive \( g_{\text{max}} \), both known to preserve low-frequency consistency. While our main experiments adopt \( g_{\text{constant}} \), Table~\ref{ab} shows that \( g_{\text{max}} \) achieves improved results in many cases. This suggests that carefully selecting or adapting the noise schedule can further enhance the generative model's ability to capture long-term dependencies and sharp transitions.

\subsection{Training Strategy of Compositional Priors.}
We analyze four training strategies for compositional priors on ETTh1 (Table~\ref{ab2}):
\begin{itemize}
    \item \( f_{\theta} \): jointly optimized with the SB model using only the bridge loss (default setting).
    \item \( f_{\theta_{\text{frozen}}} \): priors are pretrained and frozen during SB training.
    \item \( f_{\theta_{\text{MSE}}} \): priors are additionally trained with an auxiliary MSE loss to the ground truth.
    \item \( f_{\theta_{\text{MSE-N}}} \): same as above, but with randomly initialized priors.
\end{itemize}
We observe that the default setting (\( f_{\theta} \)) consistently yields the best results. The MSE-supervised variants perform worse, indicating that conflicting supervision signals may disrupt alignment with the bridge dynamics. Notably, \( f_{\theta_{\text{MSE-N}}} \) performs the worst, highlighting the necessity of prior pretraining for effective and stable optimization.

\subsection{Training Dynamics of Prior Initialization.}
To study the impact of priors on learning complex temporal patterns over time under three settings on ETTh1 with 12.5\% masking (Table~\ref{ab3}): randomly initialized priors (\( f_{\theta-N} \)), frozen pretrained priors (\( f_{\theta_{\text{frozen}}} \)), and trainable pretrained priors (\( f_{\theta} \)). Results show that pretrained and trainable priors (\( f_{\theta} \)) accelerate convergence in early training. This confirms the dual benefit of using informative, well-initialized priors: improved performance and faster optimization.

\section{Conclusion}

In this work, we introduced \textbf{Bridge-TS}, a novel generative framework for time series imputation that bridges expert deterministic predictions with probabilistic refinement via the Schrödinger Bridge. By leveraging informative priors—either from a single expert model or as a composition of multiple pretrained estimators, Bridge-TS effectively initiates the generative trajectory closer to the ground truth and performs data-to-data evolution for accurate reconstruction. Extensive experiments across six benchmark datasets and multiple missing ratios demonstrate that Bridge-TS consistently outperforms state-of-the-art deterministic and generative baselines. Furthermore, our ablation studies confirm that both the informativeness and the training strategy of expert priors critically influence performance. We believe this work highlights a promising direction in marrying deterministic estimation with principled generative modeling, opening new possibilities for robust and efficient time series imputation.
\newpage
\bibliographystyle{IEEEbib}
\bibliography{strings,refs}

\appendix

\end{document}